%% file: main.tex
\documentclass{article}
\pdfoutput=1 

\usepackage{PRIMEarxiv}

\newgeometry{
    textheight=8in,
    top=2cm,
    bottom=4cm,
  }

\usepackage[utf8]{inputenc} 
\usepackage[T1]{fontenc}    
\usepackage[hidelinks]{hyperref}

\usepackage{url}            
\usepackage{booktabs}       
\usepackage{amsfonts}       
\usepackage{nicefrac}       
\usepackage{microtype}      
\usepackage{lipsum}
\usepackage{fancyhdr}       
\usepackage{graphicx}       

\input{others/_packages.tex}

\input{others/_commands.tex}

\input{others/_metadata.tex}

\begin{document}

\maketitle
\thispagestyle{fancy}

\input{others/abstract.tex}

\input{others/body.tex}

  
\printbibliography

\end{document}

%% file: others/_packages.tex
\usepackage{amsmath,amssymb,amsfonts}
\usepackage{graphicx}
\usepackage{textcomp}
\usepackage[dvipsnames]{xcolor}
\def\BibTeX{{\rm B\kern-.05em{\sc i\kern-.025em b}\kern-.08em
    T\kern-.1667em\lower.7ex\hbox{E}\kern-.125emX}}

\usepackage{longtable} 

\newcommand{\citeauthorin}[1]{{\citeauthor{#1}~\autocite{#1}}}
\usepackage{pgfplots}
\pgfplotsset{compat=1.14}
\usepackage{subfig}
\usepackage[justification=centering]{caption}

\usepackage{float}
\restylefloat{table}
\usepackage{placeins}
\usepackage{multirow}

\usepackage[backend=bibtex, sorting=none, style=ieee]{biblatex}


%% file: others/_commands.tex
\newcommand{\tss}[1]{\textsuperscript{#1}}


%% file: others/_metadata.tex
\title{Low Complexity Approaches for End-to-End Latency Prediction
}

\author{
  Pierre Larrenie \\
  Thales SIX \& LIGM \\
  Université Gustave Eiffel, CNRS \\
  Marne-la-Vallée, France\\
  \texttt{pierre.larrenie@esiee.fr} \\
   \And
  Jean-François Bercher \\
  LIGM \\
  Université Gustave Eiffel, CNRS \\
  Marne-la-Vallée, France\\
  \texttt{jean-francois.bercher@esiee.fr} \\
  \AND
  Olivier Venard \\
  ESYCOM \\
  Université Gustave Eiffel, CNRS \\
  Marne-la-Vallée, France\\
  \texttt{olivier.venard@esiee.fr} \\
  \And
   Iyad Lahsen-Cherif \\
  Institut National des Postes et Télécommunications (INPT) \\
  Rabat, Morocco \\
  \texttt{lahsencherif@inpt.ac.ma} \\
}

%% file: others/abstract.tex
\begin{abstract}
Software Defined Networks have opened the door to statistical and
AI-based techniques to improve efficiency of networking. Especially to
ensure a certain \emph{Quality of Service} (QoS) for specific
applications by routing packets with awareness on content nature (VoIP,
video, files, etc.) and its needs (latency, bandwidth, etc.) to
use efficiently resources of a network.

Predicting various Key Performance Indicators (KPIs) at any level may
handle such problems while preserving network bandwidth.

The question addressed in this work is the design of efficient and
low-cost algorithms for KPI prediction, implementable at the
local level. We focus on end-to-end latency prediction, for
which we illustrate our approaches and results on a public
dataset from the recent international challenge on GNN \autocite{suarez2021graph}. 
We propose several low complexity, locally implementable approaches, achieving significantly lower wall time both for training and inference, with marginally worse prediction accuracy compared to state-of-the-art global GNN solutions. 
\end{abstract}

\keywords{  KPI Prediction \and Machine Learning \and General Regression \and SDN \and Networking \and Queuing Theory \and GNN}

%% file: others/body.tex
\section{Introduction}

Routing while ensuring Quality of Service (QoS) is still a great
challenge in any networks. Having powerful ways to transmit data is not
sufficient, we must use resources wisely. This is true for wide static
networks but even more for mobile networks with dynamic topology. 

The emergence of Software-Defined Networking (SDN) \autocites{singh2017sdnsurvey}{amin2018sdnsurvey} has made it possible to share data more efficiently between communication layers.
Services are able to provide network requirements to routers based on their nature; routers acquire data about network performance, and finally allocate resources to meet these requirements.
However, acquiring overall network performance can result in high consumption of network bandwidth for signalization; that is particularly constraining for networks with limited resources like Mobile \textit{Ad-Hoc} Networks (MANET).

We consider network for which we wish to reduce the amount of signalization and perform intelligent routing. In order to limit signalization, a first axis is to be able to estimate some key performance indicators (KPI) from other KPIs. A second point would be to be able to perform this prediction locally, at the node level, rather than a global estimation of the network. Finally, if predictions are to be performed locally, the complexity of the algorithms will need to be low, but still preserve good prediction quality.  
The question we address is thus the design of efficient and low-cost algorithms for KPI prediction, implementable at the local level. We focus on end-to-end latency prediction, for which we illustrate our approaches and results on a public dataset from the recent international challenge \autocite{suarez2021graph}. 

The best performances of the state-of-the-art are obtained with Graph Neural Networks (GNNs) \autocites{rusek2019unveiling}{itubnngnn2020}{suarez2021graph}. Although this is a global method while we favor local methods, we use these performances as a benchmark. 
We first propose to use standard machine learning regression methods, for which we show that a careful feature engineering and feature selection (based on queue theory and the approach in \autocite{parana2021}) allows to obtain near state-of-the-art performances with a very low number of parameters and very low computational cost, with the ability to operate at the link level instead of a whole-graph level. Building on that, we show that it is even possible to obtain similar performances with a single feature and 
 curve-fitting methods.

The presentation is structured as follows. In \autoref{sec:key_concepts}, we first recall the key concepts on GNNs and queues; present some related works in the literature, before introducing the dataset used for the validation of our proposals.  In \autoref{sec:our_approaches}, we present the different approaches proposed, starting with the choice of features for machine learning methods, followed by general curve fitting methods. We then compare in \autoref{sec:comparison_analysis} the performances of these different approaches, in terms of performance as well as in terms of learning time and inference time. Finally, we conclude, discuss the overall results and draw some perspectives.

\section{\texorpdfstring{Related work and dataset
\label{sec:key_concepts}}{Related work and dataset}}

\subsection{Graph Neural Networks (GNNs)}

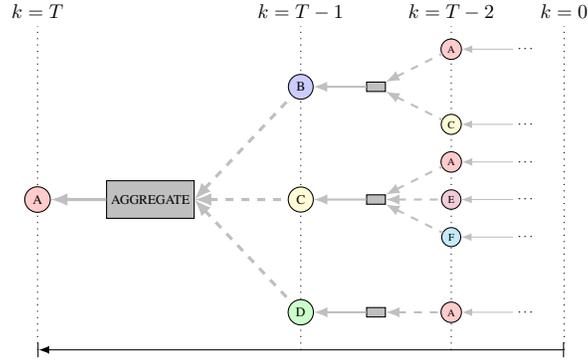
\begin{figure}
    \centering
    \input{./share/tikz/message_passing.tex}
    \caption{GNN repeating $T$ Message Passing mechanisms: message propagation and aggregation (\emph{inspired from \autocite{hamilton2020graph}})}
    \label{fig:gnn_mecanics}
    
\end{figure}

GNN
\autocites{hamilton2020graph}{bacciu2020gentle} is a machine learning
paradigm that handles non-euclidean data: graphs. A graph is defined as
a set of nodes and edges with some properties on its nodes and its
edges. The key point in GNNs is the concept of Message Passing: each
node of the graph will update its state according to states of its
neighborhood by sending and receiving \emph{messages} transmitted along
edges. By repeating this mechanism \(T\) times, a node is able to
capture states of its \(T\)-hop neighborhood as shown in
\autoref{fig:gnn_mecanics}.

\subsection{Queue Theory} 

Queue Theory is a well studied domain and for most of simple queue systems, explicit equations exist \autocite{cooper1981introduction}. Further, we will refer to queue systems by using their Kendall's notation. We often take at reference $M/M/1$ and $M/M/1/K$ for their markovian property, since equations are particularly easy to handle in this case. 

However, for more general queue systems such as $M/G/1$ and $M/G/1/K$, equations are getting more complex. Whereas closed formulas exist for $M/G/1$ queues, $M/G/1/K$ queues require to solve an equation system with $K+1$ unknowns.

Queue systems analysis focus on stable queue, i.e. when the ratio $\rho=\frac{\lambda}{\mu} \leq 1$ where $\lambda$ (resp. $\mu$) is the expected value of the arrival rate process (resp. service time). But finite queue systems are always stable since the maximal number of pending items is always finite and are subject to loss instead. To model the drop of incoming item in the queue we use the ratio $\rho_e=\frac{\lambda_e}{\mu}$ where $\lambda_e$ is known as the effective arrival rate and can be determined thanks to equation \eqref{eq:lamdae}.

\begin{equation}
\lambda_e = \lambda(1-\pi_K) = \mu(1-\pi_0) \label{eq:lamdae}
\end{equation}

Where $\pi_0$ (resp. $\pi_K$) in the above equation \eqref{eq:lamdae} refers to the probability to the queue at equilibrium to be empty (resp. full).

\subsection{Related Work}
\citeauthorin{chua2016stringer} present an heuristic and an Mixed
Integer Programming approach to optimize Service Functions Chain
provisioning when using Network Functions Virtualization for a service
provider. Their approach relies on minimizing a trade-off between the
expected latency and infrastructures resources.

Such optimization routing flow in SDN may need additional information
to be exchanged between the nodes of a network. This results in an
increase of the volume of signalization, by performing some measurements
such as in \autocite{pasca2017amps}. This is not a consequent problem in
unconstrained networks, i.e.~static wired networks with near-infinite
bandwidth but may decrease performance of wireless network with poor
capacity. An interesting solution to save bandwidth would be to predict
some of the KPIs from other KPIs and data exchanged globally between
nodes.

In \autocites{poularakis2018sdn}{poularakis2019tacticalsdn}, authors proposed a MANETs application of SDN in the domain of tactical networks. They proposed a multi-level SDN controllers architecture to build both secure and resilient networking. While orchestrating communication efficiently under military constraints such as: high-level of dynamism, frequent network failures, resources-limited devices.  The proposed architecture is a trade-off between traditional centralized architecture of SDN and a decentralized architecture to meet dynamic in-network constraints. 

\citeauthorin{jahromi2018towards} proposed a Quality of Experience (QoE) management strategy in a SDN to optimize the loading time of all the tile of a mapping application. They have shown the impact of several KPIs on their application using a Generalized Linear Model (GLM). This mechanism make the application aware of the current network state.

Promising works rely on estimating KPIs at a graph-level. Note that it is very difficult, if not impossible, to address this analytically since computer networks models a complex structure of chained interfering queues for each flow in the network.

\begin{figure}[!ht]
    \centering
    \includegraphics[width=2.5in]{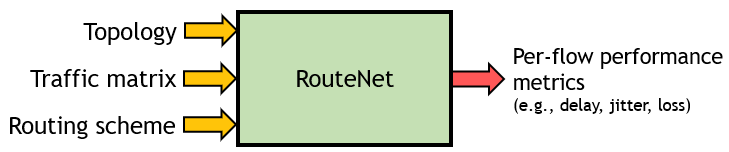}
    \caption{Routenet Architecture \autocite{rusek2019unveiling}}
    \label{fig:routenet}
    
\end{figure}

\begin{figure*}\centering
    \subfloat[Simple topology]{\input{./share/tikz/sample_topology.tex}}\quad
    \subfloat[Paths-links Hypergraph of (a)]{\input{./share/tikz/paths_links_graph.tex}}
    \caption{
        Routenet \autocite{rusek2019unveiling} paths-links hypergraph transformation applied on a simple topology graph carrying 3 flows. \\
        {\small (a) Black circles represents communication node, double headed arrows between them denotes available symmetric communications links and dotted arrows shows flows path. (b) Circle (resp. dotted) represents links (resp. flows) entities defined in the first graph ($L_{ij}$ is the symmetric link between node $i$ and node $j$.). Unidirectional arrows encode the relation "\textless flow\textgreater~is carried by \textless link\textgreater ".}
    }
    \label{fig:routenet_hypergraph}

\end{figure*}
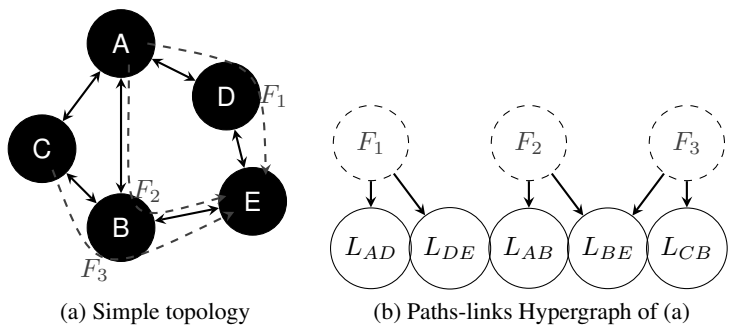

\citeauthorin{rusek2019unveiling} used GNNs for predicting
KPIs such as latency, error-rate and jitter. They relied on the
\emph{Routenet} architecture of \autoref{fig:routenet}. The idea is to
model the problem as a bipartite hypergraph mapping flows to links as
depicted on \autoref{fig:routenet_hypergraph}. Aggregating messages in
such graph may result in predicting KPIs of the network in input. The
model needs to know the routing scheme, traffic and links properties.
Their result is very promising and has been the subject of two ITU
Challenge in 2020 and 2021 \autocites{itubnngnn2020}{suarez2021graph}.
These ITU challenges have very good results since the top-3 teams are around
2\% error in delay prediction in the sense of Mean-Absolute Percentage Error (MAPE).

In \autocite{parana2021}, very promising
results were obtained with a a near 1\% GNN model error (in the sense of MAPE)
on the test set.
The model mix analytical \(M/M/1/K\) queueing theory used to create
extra-features to feed GNN model. In order to satisfy the constraint of
scalability proposed by the challenge, the first part of model operates
at the link level.

\subsection{\texorpdfstring{Dataset
\label{sec:dataset}}{Dataset }}

\begin{figure}[!ht]
    \centering
    \includegraphics[width=3in]{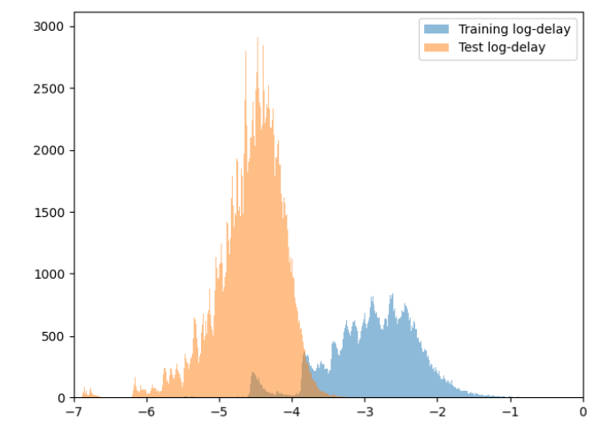}

    \caption{End-to-end latency distribution on train and test datasets of ITU Challenge 2021, where train and test datasets describe networks of very different sizes.\autocite{suarez2021graph}}
    \label{fig:dataset_delay_dist}
    
\end{figure}

We use public data from the challenge \autocite{suarez2021graph}
The dataset models static networks that have run for a certain amount of time; the obtained data
is a mean of the global working period. The data contains information
about the topology of the network, participants and available link
characteristics, traffic and routing information. The aim of the GNN ITU
Challenge \autocite{suarez2021graph} was to build a scalable GNN model in
order to predict end-to-end flow latency. Nevertheless, train on one
hand and test and validation on the other hand model very different
networks. Whereas training dataset models network between 25 and 50
nodes (120,000 samples), test (1,560 samples) and validation (3,120
samples) datasets model networks up to 300 nodes. This results in a very
different distribution among these different splits as shown on
\autoref{fig:dataset_delay_dist}.

It is important to point out that the proposed data is not in accordance
with \(M/M/1/K\) queue models since process service time depends on the
size of the packet. The size of the packet for each flow follows a
Binomial distribution; it can be approximated by a Normal distribution
inducing a general service time.

Nevertheless, it turns out that the system does not have the behavior of
a \(M/G/1/K\) queue system globally but that of a complex system with
interconnected queues that cannot be easily modeled.

Hence, approximating the system locally by a mixed of a simple
analytical theory (\(M/M/1/K\)) and black-box optimization (GNNs), as was proposed in \autocite{parana2021},  is a
good approach despite the lack of explicability or interpretability
and the high-computational requirements with a lot of parameters to
train. We show below that it is possible to obtain comparable performances with other regression approaches.

\section{\texorpdfstring{Our approaches
\label{sec:our_approaches}}{Our approaches}}

The main question is to define an estimator $\hat{y}$ of the occupancy $y$ according to the various available characteristics of the system, with a joint objective of low complexity and performance. In the following, we present regression approaches based on machine learning and then approaches based on curve-fitting. 

Once an estimate of occupancy is obtained, it is possible to get the latency prediction $\hat{d_n}$ for a specific link $n$ by the simple relation
\begin{equation*}
    \hat{d_n} = \hat{y_n}\frac{\mathbb{E}(|P_n|)}{c_n}
\end{equation*}
where $\mathbb{E}(|P_n|)$ is the observed average packet size on link $n$ and $c_n$ the capacity of this link.

Performances will be evaluated using the MAPE loss-function
\begin{equation}
    \mathcal{L}\left(\hat{y}, y \right) = \frac{100\%}{N} \sum_{n=1}^N \left|\frac{\hat{y}_n - y_n}{y_n}\right|
\end{equation}
which is preferred to Mean Squared Error (MSE) because of its scale-invariant property.

\subsection{\texorpdfstring{Feature Engineering and Machine Learning
\label{sec:feature_engineering}}{Feature Engineering and Machine Learning}}

Based on the assumption that the system may be approximated by a model whose essential features come from \(M/M/1/K\) and \(M/G/1/K\) queue theory,

we took essential parameters characterizing queueing systems, such as: $\rho$, $\rho_e$, $\pi_0$, $\pi_K$, etc. and built further features by applying interactions and various non-linearities (powers, log, exponential, square root). Then, we selected features in this set by a forward step-wise selection method; i.e. by adding in turn each feature to potential models and keeping the feature with best performance. Finally, we selected the model with best MAPE error. For a linear regression model, this led us to select and keep a set of 4 simple features, which interestingly enough, have simple interpretations:

\begin{equation}
    \begin{cases}
        \pi_0 = \frac{1-\rho}{1-\rho^{K+1}} \\
        L = \rho + \pi_0\sum_k k\rho^k \\
        \rho_e = \frac{\lambda_e}{\lambda}\rho = \frac{\lambda_e}{\mu} \\
        S_e = \sum_k k\rho_e^k
    \end{cases}\label{eq:mvfeatures}
\end{equation}
where $L$ is the expected number of packets in the queue according to $M/M/1/K$, $\pi_0$ the probability that the queue is empty according to $M/M/1/K$ theory, $\rho_e$ the effective queue utilization, and $S_e$ the unnormalized expected value of the effective number of packet in the queue buffer.
These features can be thought as a kind of data preprocessing, before applying ML algorithms, and this turns out to be a key to achieving good performances. The 4 previous features have been kept as input for all the machine learning models.  

Next we considered several machine learning algorithm, fitted on the training split and performances were evaluated by test split of a public dataset \autocite{suarez2021graph}.
Algorithms that were considered are: Multi-Layer Perceptron model (MLP) with 4 layers and with ReLU activation function,
Linear Regression, Gradient Boosting Regression Tree (GBRT) with an
ensemble of \(n=100\) estimators, Random Forest of \(n=100\) trees and
Generalized Linear Model (GLM) with Poisson family and exponential link.
All results of these methods are shown in \autoref{tab:benchmark}.

\subsection{Curve Regression for occupancy prediction}
\label{sec:curveregression}

\begin{figure}[!ht]
    \centering
    \includegraphics[width=3in]{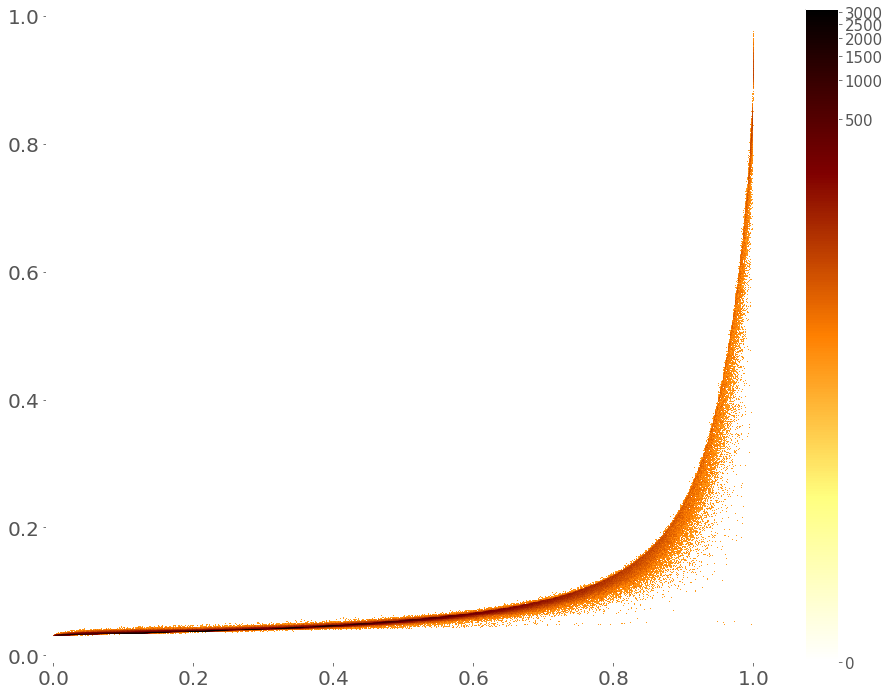}

    \caption{Data of ITU Challenge 2021 \autocite{suarez2021graph}, $\rho_e$ vs queue occupancy. Color-scale is an indicator of points cloud density.}
    \label{fig:datarhoy}
    
\end{figure}

There is a high interdependence of the features we selected in 
\autoref{eq:mvfeatures}, since all these features can be expressed in term of \(\rho_e\).
Furthermore, it is confirmed by data exploration that $\rho_e$ is the prominent feature for occupancy prediction (and in turn latency prediction), as exemplified in \autoref{fig:datarhoy}. 

It is then tempting to try to further simplify our features space and to try to estimate the occupancy from a non-linear transformation of the single feature $\rho_e$, as:
\begin{equation}
    \hat{y} = g(\rho_e)
\end{equation} 
where \(\hat{y}\) is the estimate of the occupancy \(y\). 
The concerns are of course to define simple and efficient functions $g$, with a low number of parameters, that can model the  kind of growth shown in \autoref{fig:datarhoy}, and of course to check that the performance remains interesting.

We followed three approaches to design the estimator $g$ in order to predict links occupancy and end-to-end flow latency. In all cases, the parameters of $g$ were computed by minimizing the mean squared of the regression error.

\subsubsection{Exponential of polynomial}

The simplest approach is to use a curve-fitting regression of the form
\begin{equation}
    \hat{y}=g(\rho_e)=e^{p_n(\rho_e)}
\end{equation}
where \(p_n(x)\in \mathbb{R}_n[x]\) is a polynomial of degree \(n\) with real coefficients. 

In order to find coefficients of \(p_n\) one can obviously consider
predicting \(\log(y)\) (where \(y\) denotes the queue occupancy).
Choosing an arbitrary high polynomial degree results to oscillations
and increases largely computation time. However choosing a too small
degree does not allow the prediction of high occupancy.

\subsubsection{Generative polynomials}

The estimator $g$ is defined as a linear combination of simple functions \((f_n)\):
\begin{equation}
    \hat{y}=g(\rho_e)=\sum_n \alpha_n \cdot f_n(\rho_e)
\end{equation}

\paragraph{\texorpdfstring{Generative polynomial similar to \(M/M/1/K\)
theory}{Generative polynomial similar to M/M/1/K theory}}

The idea here is to use a polynomial $f^K_n$ that will match approximately the expression in  \autoref{eq:mvfeatures} of the expected number of packets in the queue  \(L\). 
\begin{equation}
    \begin{cases}
        f^K_n(x)=\frac{\phi^K_n(x)}{\gamma_n}\\
        \phi^K_n(x)=x^n\frac{(1-x)}{1-x^{K+1}} & n\geq 0 \; \forall x\in[0;1[\\
        \gamma_n = \phi^K_n\left(\frac{n}{n+1}\right) & n\geq 0
    \end{cases}
\end{equation}
where $K$ is the size of the queue\footnote{In results shown Table I., we consider $K=32$ in order to match the data contained in the ITU challenge dataset \autocite{suarez2021graph}.} 
The sequence of \((f_n)_{n=0}^{K}\) is finite and defined in interval \([0;1[\).

In order to improve regression capabilities, each $f_n^K$ is defined as $\phi^K_n$  normalized by \(\gamma_n\), a local maximum of \(\phi^K_n\) in the interval \([0;1[\).

\paragraph{Bernstein Polynomials}

The previous method relies on polynomial approximation. Since the expected value $L$ can be expressed theoretically in terms of a polynomial
of degree \(K\), we are driven to the Bernstein polynomials that form a basis in the set of polynomial in the interval \([0;1]\):
\begin{equation}
    f^K_n(x) = \binom{K}{n} x^n(1-x)^{K-n}
\end{equation}
The approximation of any continuous function on \([0;1[\) by a Bernstein polynomial converges uniformly.

\subsubsection{Implicit function}

The idea here is to define a set of \(N\) points
\(\theta_n=(a_n, b_n)\) and approximate the underlying function by linear interpolation between those points.  To obtain a good positioning of these points, we select them as the solution of the following optimization problem:

\begin{equation}
    \begin{aligned}
        \min_{\theta} \quad & \mathcal{L}(f_\theta(x), y) + \frac{\alpha}{N} \sum_n \frac{\| \vec{u}_n \times \vec{u}_{n+1} \|^2}{\|\vec{u}_n\|^2\|\vec{u}_{n+1}\|^2}\\
        \textrm{s.t.} \quad & \vec{u}_n = \theta_{n+1} - \theta_0n\\
        & a_0 = 0 \\
        & a_N = b_N = 1 \\
        & a_{n+1} - a_n \geq 0    \\
        & \theta_n = (a_n,\, b_n)^T \in [0; 1]^2
    \end{aligned}
    \label{eq:implicit_reg}
\end{equation}

\autoref{eq:implicit_reg} includes a first term for minimizing the interpolation error, and a second term weighted by a parameter \(\alpha\geq0\), to force
\(\theta_n\) sequence to be as aligned and as far as possible. This
implies that our sampling will be refined in high curvature zone of our
function. The constraint formulated makes \(\theta_n\) an increasing sequence along the feature axis in order to get a correct interpolation of the curve, especially when \(N\) is high enough.

\section{\texorpdfstring{Comparison and Discussion
\label{sec:comparison_analysis}}{Comparison and Discussion }}

\begin{table*}[thp]
\caption{
    Results Synthesis of various models for flow latency prediction. Test dataset from \autocite{suarez2021graph}\\
      {\small \tss{*}only 500,000 samples used for training (2.25\% of training dataset); \tss{**}only 5,000,000 samples used for training (22.5\% of training dataset); \tss{+}under-estimation/over-estimation occurs on high queue occupancy prediction}
}
\label{tab:benchmark}
\centering
\input{share/tab/bench}
\end{table*}

In this section, we evaluate our methods on the data from the GNN ITU Challenge 2021, described in \autoref{sec:dataset}. We compare our results to those of the challenge winners, which establish the state-of-the-art in terms of pure performance. Since the actual labeled test dataset used for the challenge was released after the end of the challenge, all evaluations are performed on this particular dataset. 
The \autoref{tab:benchmark} presents the characteristics of the methods, in terms of the number of input features and parameters to be learned; their performance in the sense of MAPE and MSE; and the values of the execution times, both in learning time and inference time. All results were obtained with the same computer configuration:  120 Go RAM, 1 CPU Intel i9-9920X @ 3.50 GHz with 24 cores and 2 GPUs Nvidia TITAN RTX2080 24Go.  

The methods used for comparison are divided into 3 groups, the first being the set of GNN approaches.

In the second group, we used classical machine learning models with only 4 input features obtained by stepwise selection, as presented in \autoref{eq:mvfeatures}.

In the third group, we group curve regression models using a single well-chosen feature, namely \(\rho_e\), as presented in \autoref{sec:curveregression}.

As we can observe, the proposed approaches achieve a much lower computational time than the GNN approaches, both in terms of learning time and inference time; this at the cost of a marginal performance degradation. 

Moreover, non-GNN approaches provide a more local solution since predictions are performed at the link level and not at the whole graph level (Models predict queues occupancy, then compute analytically delay for each link and finally aggregate along path). This would allow to use them for simple local predictions, without having to rely on the global knowledge and prediction of the network. 

The consequent gain in computational time of our low-complexity approaches is that they use far fewer parameters, which reduces the amount of data needed for training. The reduction in the number of parameters and the architecture (number of operations) of the solutions explains the drop in learning and inference times. 

Nevertheless, when we match the distribution as presented in
\autoref{fig:datarhoy}, we notice that most of our data are on a low occupancy level. In practice, some models have a kind of limited behavior when the occupation of the targeted queue is close to 100\%: there is a significant over- or under-prediction. However, this behavior does not really affect the overall performance due to the low density of this scenario in our dataset and the predicted values are close enough to the targets.

\section{Conclusion}

In this paper, we considered the problem of designing efficient and low-cost algorithms for KPI prediction, implementable at the local level. 
We have argued and proposed several alternatives to GNNs for predicting the queue occupancy of a complex system using simple ML models with carefully chosen features or general curve-fitting methods. 

At the cost of a marginal performance loss, our proposals are characterized by low complexity, significantly lower learning and inference times compared to GNNs, and the possibility of local deployment.
Thus, this type of solution can be used for continuous performance monitoring. 

The low complexity and structures of linear regression algorithms or curve-fitting solutions should also be suitable for adaptive formulations.  These last two points are current perspectives of this work. Of course, the approaches considered here will have to be considered and adapted for other types of KPI, such as error-rate or jitter.  

Finally, a last point that deserves interest is the fact that these low complexity models can be interpreted/explained either by direct inspection (visualization), or by using tools such as Shapley values \autocite{lundberg2017unified} which allow to interpret output values by measuring contributions of each input feature on the prediction.

%% file: share/tikz/message_passing.tex
\begin{tikzpicture}
    \begin{scope} 
        \draw[|<-|, >=latex] (0,0) -- (7,0);
        \draw[dotted] (0, 0) -- (0,4.5) node[rectangle, fill=white, scale=0.75] {$k=T$};
        \draw[dotted] (3.5, 0) -- (3.5,4.5) node[rectangle, fill=white, scale=0.75] {$k=T-1$};
        \draw[dotted] (5.5, 0) -- (5.5,4.5) node[rectangle, fill=white, scale=0.75] {$k=T-2$};
        \draw[dotted] (7, 0) -- (7,4.5) node[rectangle, fill=white, scale=0.75] {$k=0$};

    \end{scope}

    \begin{scope}[shift={(0,2)}, every node/.style={circle,draw, scale=0.5}] 
        \node [fill=red!20] (A) at (0,0) {A};
        \node (AGG)  [rectangle, fill=gray!50, minimum height=1cm] at (1.5,0) {AGGREGATE};
        \node [fill=blue!20] (B) at (3.5,1.5) {B};
        \node [fill=yellow!20] (C) at (3.5,0) {C};
        \node [fill=green!20] (D) at (3.5,-1.5) {D};
     
        \node (AGG_B_T2)  [rectangle, fill=gray!50, minimum height=0.25cm, minimum width=0.5cm] at (4.5,1.5) {};   
        \node (B2_A) [scale=0.8, fill=red!20] at (5.5, 2) {A};
        \node (B2_C) [scale=0.8, fill=yellow!20] at (5.5, 1) {C};Download

        \node (AGG_C_T2)  [rectangle, fill=gray!50, minimum height=0.25cm, minimum width=0.5cm] at (4.5,0) {};
        \node (C2_A) [scale=0.8, fill=red!20] at (5.5, 0.5) {A};
        \node (C2_E) [scale=0.8, fill=purple!20] at (5.5, 0) {E};
        \node (C2_F) [scale=0.8, fill=cyan!20] at (5.5, -0.5) {F};
        
        \node (AGG_D_T2)  [rectangle, fill=gray!50, minimum height=0.25cm, minimum width=0.5cm] at (4.5,-1.5) {};
        \node (D2_A) [scale=0.8, fill=red!20] at (5.5, -1.5) {A};
        
        \node (AGG_BA_T3)  [rectangle, fill=white, draw=none] at (6.5, 2) {$\cdots$};
        \node (AGG_BC_T3)  [rectangle, fill=white, draw=none] at (6.5,1) {$\cdots$};
        
        \node (AGG_CA_T3)  [rectangle, fill=white, draw=none] at (6.5, 0.5) {$\cdots$};
        \node (AGG_CE_T3)  [rectangle, fill=white, draw=none] at (6.5,0) {$\cdots$};
        \node (AGG_CF_T3)  [rectangle, fill=white, draw=none] at (6.5,-0.5) {$\cdots$};

        \node (AGG_DA_T3)  [rectangle, fill=white, draw=none] at (6.5, -1.5) {$\cdots$};
    \end{scope}

    \begin{scope}[<-,>=latex, 
                  every edge/.style={draw=gray!50},
                  ]
        \path[very thick] (A) edge (AGG) ;
        \path[thick] (B) edge (AGG_B_T2);
        \path[thick] (C) edge (AGG_C_T2);
        \path[thick] (D) edge (AGG_D_T2);
        
        \path (B2_A) edge (AGG_BA_T3);
        \path (B2_C) edge (AGG_BC_T3);
        \path (C2_A) edge (AGG_CA_T3);
        \path (C2_E) edge (AGG_CE_T3);
        \path (C2_F) edge (AGG_CF_T3);
        \path (D2_A) edge (AGG_DA_T3);

    \end{scope}
    
    \begin{scope}[<-,>=latex, 
                  every edge/.style={draw=gray!50, dashed},
                  ]
        \path[very thick] (AGG.east)    edge (B)
                                        edge (C)
                                        edge (D);
        
        \path[thick] (AGG_B_T2) edge (B2_A)
                                edge (B2_C);
                                
        \path[thick] (AGG_C_T2) edge (C2_A)
                                edge (C2_E)
                                edge (C2_F);

        \path[thick] (AGG_D_T2) edge (D2_A);
        
    \end{scope}
\end{tikzpicture}

%% file: share/tikz/sample_topology.tex
\begin{tikzpicture}[scale=0.7]
        \begin{scope}[node_topo/.style={circle,fill,draw,text=white,font=\sffamily,minimum
        size=9mm}, edge_topo/.style={thick, stealth-stealth}, flow_topo/.style={darkgray, -stealth, thick, dashed}]
            \node[node_topo] (v1) at (0.5,1) {A};
            \node[node_topo](v4) at (0.5,-2.5) {B};
            \node[node_topo] (v3) at (-1,-1) {C};
            \node[node_topo] (v2) at (2.5,0) {D};
            \node[node_topo] (v5) at (3,-2) {E};
            \draw[thick, stealth-stealth]  (v1) edge (v2);
            \draw[edge_topo]  (v1) edge (v3);
            \draw[edge_topo]  (v4) edge (v3);
            \draw[edge_topo]  (v4) edge (v5);
            \draw[edge_topo]  (v1) edge (v4);
            \draw[edge_topo]  (v2) edge (v5);
            \draw[flow_topo]
                        (-0.8,-1.4) .. controls (0,-3.5)  and (0,-3.5) .. (2.65,-2.2);
            \node[darkgray] (F1) at (0,-3.3) {$F_3$};
            \draw[flow_topo]
                        (0.65,0.60) .. controls (0.65,-2.5)  and (0.355,-2.5)  .. (2.5,-1.9);
            \node[darkgray] (F2) at (1,-1.8) {$F_2$};
            \draw[flow_topo]
                        (1,1) .. controls (3.2,0.6)  and (3.2,0.6)  .. (3.25,-1.5);
            \node[darkgray] (F3) at (3.4,0) {$F_1$};
        \end{scope}
\end{tikzpicture}

%% file: share/tikz/paths_links_graph.tex
\begin{tikzpicture}[scale=0.7]
\begin{scope}[flow_node/.style={circle,dashed,draw,text=darkgray,font=\sffamily,minimum
            size=10mm},link_node/.style={circle,draw,text=black,font=\sffamily,minimum size=10mm}, link_use/.style={black, -stealth, thick}]
            \node[flow_node] (F1) at (-1,0) {$F_1$};
            \node[flow_node] (F2) at (2,0) {$F_2$};
            \node[flow_node] (F3) at (5,0) {$F_3$};

            \node[link_node] (LAD) at (-1, -2) {$L_{AD}$};
            \node[link_node] (LDE) at (0.5,-2) {$L_{DE}$};
            \node[link_node] (LAB) at (2,-2) {$L_{AB}$};
            \node[link_node] (LBE) at (3.5,-2) {$L_{BE}$};
            \node[link_node] (LCB) at (5,-2) {$L_{CB}$};

            \draw[link_use]  (F1) edge (LAD)
                                edge (LDE);			
            \draw[link_use]  (F2) edge (LAB)
                                edge (LBE);
            \draw[link_use]  (F3) edge (LBE)
                                edge (LCB);
            \end{scope}
\end{tikzpicture}

%% file: share/tab/bench.tex
\resizebox{\textwidth}{!}{%
\begin{tabular}{|c|c|c|c|c|c|c|}
\hline
Approaches &
  Input Features &
  \begin{tabular}[c]{@{}c@{}}Model \\ Parameters\end{tabular} &
  MAPE &
  MSE &
  \begin{tabular}[c]{@{}c@{}}Wall \\ Training Time\end{tabular} &
  \begin{tabular}[c]{@{}c@{}}Wall \\ Inference Time\end{tabular} \\ \hline
Routenet \autocite{rusek2019unveiling} &
  \multirow{3}{*}{\begin{tabular}[c]{@{}c@{}}Topology \\ Traffic matrix\\ Routing Scheme \end{tabular}} &
  - &
  $\gg 100$\% &
  (N/A) &
  $\approx$12h & -
   \\ \cline{1-1} \cline{3-7} 
\begin{tabular}[c]{@{}c@{}}Top-1 ITU Challenge Team\\ (PARANA)\autocite{parana2021}\end{tabular} &
   &
  654006 &
  1.27\% &
  1.10e-5 &
  $\approx$8h & 214s
  \\ \hline
MLP &
  \multirow{5}{*}{\begin{tabular}[c]{@{}l@{}}$\pi_0$ \\ $L$\\  $S_e$\\  $\rho_e$\end{tabular}} &
  291 &
  1.91\% &
  3.18e-5 &
  $\approx$45min & 8.26s
   \\ \cline{1-1} \cline{3-7} 
Linear Regression\tss{*+} &
   &
  4 &
  1.74\% &
  3.20e-5 &
  \textless{}1sec & 0.296s
   \\ \cline{1-1} \cline{3-7} 
GBRT (n=100)\tss{*} &
   &
  4 &
  1.73\% &
  2.90-5 &
  $\approx$1min & 0.867s
   \\ \cline{1-1} \cline{3-7} 
Random Forest (n=100)\tss{*} &
   &
  4 &
  1.69\% &
  3.00e-5 &
  \textless{}1sec & 0.994s
   \\ \cline{1-1} \cline{3-7} 
GLM - Poisson\tss{+} &
   &
  4 &
  3.68\% &
  5.09e-4 &
  $\approx$1min &0.481s
   \\ \hline
Curve-fitting exponential (deg=3)\tss{+} &
  \multirow{6}{*}{$\rho_e$} &
  4 &
  3.94\% &
  3.75e-4 &
  $\approx$1sec & 0.311s
   \\ \cline{1-1} \cline{3-7} 
Curve-fitting exponential (deg=8)\tss{+} &
   &
  9 &
  1.70\% &
  3.53e-5 &
  $\approx$5secs & 0.320
   \\ \cline{1-1} \cline{3-7} 
Curve-fitting M/M/1/K\tss{**+} &
   &
  33 &
  2.04\% &
  4.42e-5 &
  $\approx$3min & 3.55s
   \\ \cline{1-1} \cline{3-7} 
Curve-fitting Bernstein\tss{**+} &
   &
  33 &
  1.68\% &
  3.13e-5 &
  $\approx$2min & 3.14s
   \\ \cline{1-1} \cline{3-7} 
Sampling Optimization ($N=12$, $\alpha=0$)\tss{*+} &
   &
  24 &
  1.77\% &
  3.18e-5 &
  $\approx$1min & 0.281s
   \\ \cline{1-1} \cline{3-7} 
Sampling Optimization ($N=12$, $\alpha=$1e-5)\tss{*} &
   &
  24 &
  1.77\% &
  3.18e-5 &
  
  $\approx$1min &  0.306s
   \\ \hline
\end{tabular}%
}